\documentclass[runningheads]{llncs}
\usepackage{graphicx}
\usepackage{threeparttable}
\usepackage{listings}
\usepackage{float}
\usepackage{amsmath}
\usepackage{tikz}
\usepackage{pgfplots}
\newcommand{\avsum}{\mathop{\mathpalette\avsuminner\relax}\displaylimits}

\makeatletter
\newcommand\avsuminner[2]{%
  {\sbox0{$\m@th#1\sum$}%
   \vphantom{\usebox0}%
   \ooalign{%
     \hidewidth
     \smash{\vrule height\dimexpr\ht0+1pt\relax depth\dimexpr\dp0+1pt\relax}%
     \hidewidth\cr
     $\m@th#1\sum$\cr
   }%
  }%
}
\makeatother
\begin{document}
\title{Feature-Less End-to-End Nested Term Extraction}
%

\author{Yuze Gao\inst{1} \and
Yu Yuan\inst{2}}
\authorrunning{Gao and Yu}
%
\institute{The Institute for Infocomm Research of A*STAR, Singapore 
\email{yuze.gao@outlook.com}\\ \and
School of Languages and Cultures, Nanjing University of Information Science and Technology, China\\
\email{hittle.yuan@gmail.com}}

\maketitle              
\begin{abstract}
In this paper, we proposed a deep learning-based end-to-end method on domain specified automatic term extraction (ATE),  it considers possible term spans within a fixed length in the sentence and predicts them whether they can be conceptual terms. In comparison with current ATE methods, the model supports nested term extraction and does not crucially need extra (extracted) features. Results show that it can achieve a high recall and a comparable precision on term extraction task with inputting segmented raw text.
\end{abstract}
\section{Introduction}\label{section:intro}
Automatic Term Extraction (ATE) or terminology extraction, which is to automatically extract domain specified phrases from a given corpus of a certain academic or technical domain, is widely used in text analytic like topic modelling, data mining and information retrieval from unstructured text. To specify, Table ~\ref{tab:terms} shows a simple example of the tasks, the numbers in the brackets (for example [0, 4]) indicate the start and end index of the term in the sentence separately. Given a sentence, the task is to extract the [0, 4], [0, 5], [1, 1] terms which are specific terms in a domain.
\begin{table}[]\centering
\caption{Terms Example in sentence}
\begin{tabular}{|c|l|l|}
\hline
Sentence & \multicolumn{2}{l|}{"Mouse interleukin-2 receptor alpha gene expression"}\\ \hline
{\begin{tabular}[c]{@{}c@{}}Terms \end{tabular}} 
     & \multicolumn{2}{l|}{1. {\color{red}{{[}0, 4{]}}} --\textgreater \quad "\#DNA domain or region"}\\ \cline{2-3} 
    {To Extract} & \multicolumn{2}{l|}{2. {\color{red}{[}0, 5{]}}    --\textgreater \quad "\#other\_name"}\\ \cline{2-3} 
     & \multicolumn{2}{l|}{3. {\color{red}{[}1, 1{]}} --\textgreater  \quad "\#protein\_molecule"} \\ \hline
\end{tabular}
\label{tab:terms}
\end{table}

Typically, ACE approaches make use of linguistic information (part of speech tagging, phrase chunking and constituent parsing), extracted features or defined rules to extract terminological candidates, i.e. syntactically plausible terminological noun phrases (NPs). Furthermore, in some approaches, potential terminological entries are then filtered from the candidate list using statistical, machine learning or deep learning methods. Once filtered, with low ambiguity and high specificity, these terms are particularly useful for conceptualizing a knowledge domain or for supporting the creation of a domain ontology or a terminology base.
\section{Related Works}
Current methods can mainly be divided into five kinds: rule-based method, statistical method, predominant hybrid-based, machine learning-based and deep learning-based.

Rule-based approaches~\cite{ruleace,cvalue} heavily rely on the syntax information. The portability and extensibility is also low. Error propagation from the syntax information would hamper the accuracy of models. For example, the POS-tag rule-based system~\cite{jate2} suffer from low recall due to erroneous POS-tagging. Moreover, complex structure using modifier always pose parsing challenges for most simple POS-tag rule-based algorithms.

Statistical ATE system~\cite{statiace} calls for large quality and quantity dataset to make a reasonable statistic of frequency, distribution and etc.. When predicting, the low frequency or new-occur term may be easily neglected.

Predominant hybrid ATE tries to combine the advantages of both rule-based and statistical approach. Generally, statistical methods are employed to trim the search space of candidates terms that identified by various linguistic heuristic and rules. However, combining the linguistic filters and statistical distribution ranking would lead to a degenerated precision with the increase of recall.

Machine-learning based ATE~\cite{yuyuan,ref_zhangsu,tkec2012,mlmethod} is to design and learn different features in the raw text or from syntax information, and then integrate these features into a machine learning method (such as conditional random field, supporting vector classifier). However, different domain, especially language shares different feature patterns, making this method specified to one language or domain.

Deep learning-based ATE like sequence labelling methods~\cite{seqlabel2} are also proposed recent years, but they do not support nested term extraction. ~\cite{featureless} also proposed a co-training method using CNN and LSTM, and expand its training data by adding high confidence predicted results, but this method easily leads to error propagation~\cite{errorp}.

To overcome some disadvantages of current ATE methods, we proposed another end-to-end method that based on deep learning, it supports nested term extraction and can achieve comparable experiment results without using extra features and syntax information. \\\\
To summarize, there are following contributions in this paper:

1. A novel term extraction method builds on span classification and ranking is proposed.

2. The proposed model supports nested term extraction, making it easier to do nested term extraction to form different conceptual meaning(For example, the terms nested in span [0, 5] in Tab.~\ref{tab:termexample} contains different conceptual aspect).

3. The model is a feature-free system. Apart from the segmented raw text, other various features is not a must.

4. Sentence level information are leveraged into term candidates via targeted attention mechanism, making the model more precise and interpretable.\\
The \textbf{code} and \textbf{data} is shared on github\footnote{{\url{https://github.com/CooDL/NestedTermExtraction}}}.
\section{Model}
In our model, we formulate the term extraction task as a progress of classifications and filtering, which consider all possible spans or segmentation in the sentence and distinguish them whether they can be domain specified terms in the sentence (see Fig.~\ref{fig3} for more details about our model architecture). In the following sections, we will illustrate our classification and ranking-based term extraction system in details.
\subsection{Term Spans}\label{sec:termpsans}
First, we would introduce the span or token segment (~\cite{spancr} also used in coreference resolution) used in our model.

To specify, given a sentence {\textbf{\textit{S}}=${w_1,w_2,..., w_i,...,w_n}$} with \textbf{\textit{n}} words, supposing every token sequence fragment $[w_i, w_j] (1 \leq i \leq j \leq n)$ is a span candidate, then there will be \textit{\textbf{T}} = ${\frac{\textit{\textbf{n(n+1)}}}{\textbf{\textit{2}}}}$ term span candidates in a sentence with $n$ words.

In our model, we suppose that the term maximum length is \textbf{\textit{k}}($1 \leq $\textbf{\textit{k}} $\leq \textbf{\textit{n}}$) and only consider the span candidates whose lengths are less than or equal to \textbf{\textit{k}} in the sentences. So, for each sentence with length \textbf{\textit{n}}, there finally would be ${\textit{\textbf{n*k - }}{{\frac{\textit{\textbf{k(k-1)}}}{\textit{\textbf{2}}}}}}$ term span candidates. Our task is to find the potential term spans that carry conceptual knowledge in these candidates. Refer to Tab.~\ref{tab:termexample} for more details about term span.
\begin{table}[]\centering
\caption{Term Span Example}
\begin{tabular}{|c|l|l|}
\hline
Sentence (\textbf{\textit{n}}=6) & \multicolumn{2}{l|}{"Mouse interleukin-2 receptor alpha gene expression"}\\ \hline
{\begin{tabular}[c]{@{}c@{}}True Term Spans\end{tabular}} 
     & \multicolumn{2}{l|}{{[}0, 4{]}, {\color{blue}{[}0, 5{]}}, {[}1, 1{]}}\\ \cline{2-3}  \hline
\begin{tabular}[c]{@{}c@{}}Model Processed \\ Spans (\textbf{\textit{k}}=5)\end{tabular} 
    & \multicolumn{2}{l|}{\begin{tabular}[c]{@{}l@{}}
    {[}0, 0{]}, {[}0, 1{]}, {[}0, 2{]}, {[}0, 3{]}, {\color{red}{[}0, 4{]}}, {\color{red}{[}1, 1{]}}, {[}1, 2{]}, {[}1, 3{]}, \\
    {[}1, 4{]}, {[}1, 5{]}, {[}2, 2{]}, {[}2, 3{]}, {[}2, 4{]}, {[}2, 5{]}, {[}3, 3{]}, {[}3, 4{]},\\
    {[}3, 5{]}, {[}4, 4{]}, {[}4, 5{]}, {[}5, 5{]}\end{tabular}} \\ \hline
\end{tabular}
\label{tab:termexample}
\end{table}
\subsection{Model Architecture}
\begin{figure}
\includegraphics[width=110mm,scale=0.7]{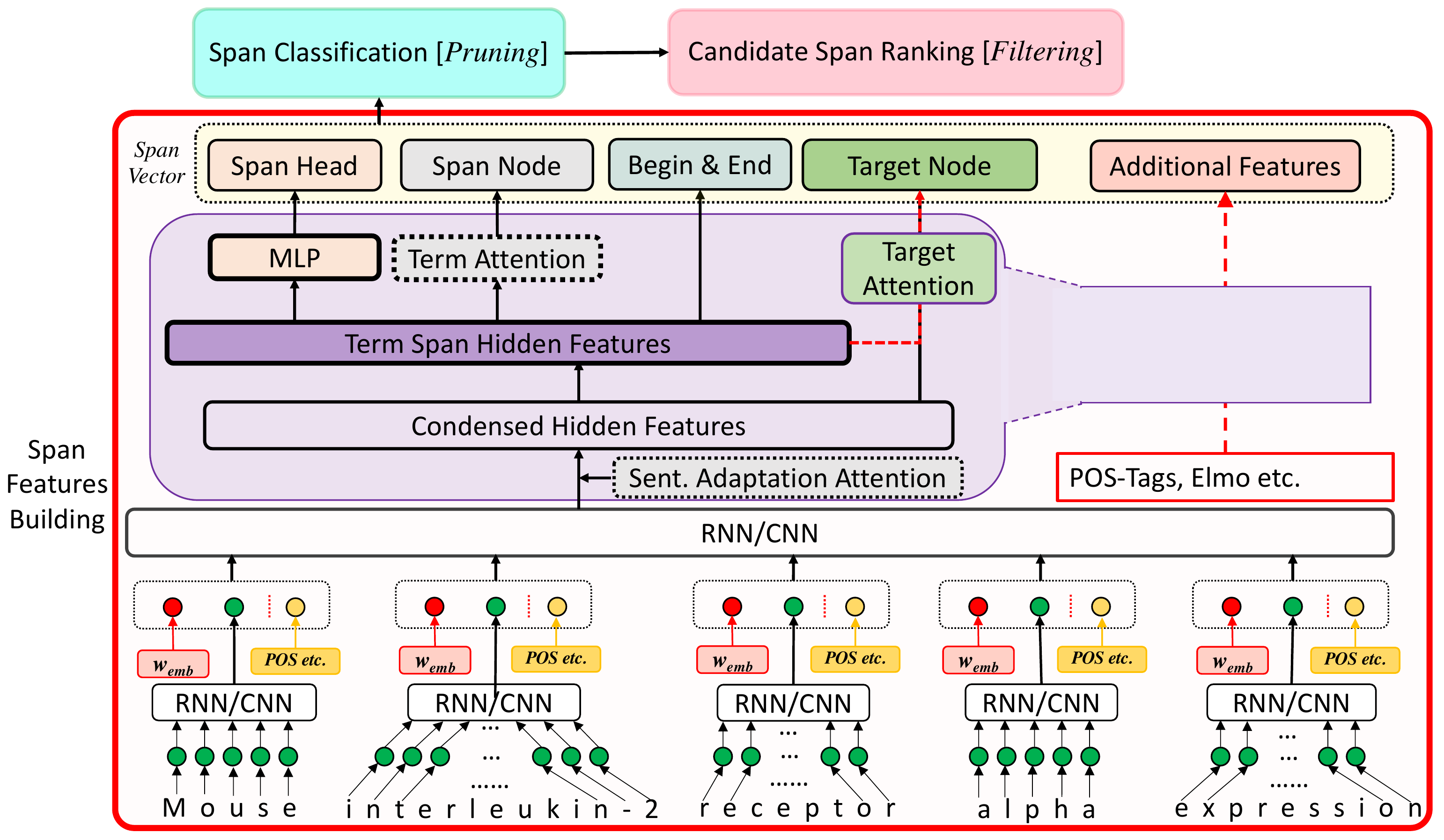}\centering
\caption{Model Architecture} \label{fig3}
\end{figure}

Briefly, our model can be divided into three parts (see Fig.~\ref{fig3} for more details): First is the feature preparation part (in red rectangle) that builds span representation vectors. Second is the classification part that classifies the span candidates to collect 'true positive spans' (TPS, that are potential to be terms). Finally is the ranking part that ranks the collected TPS based on ranking scores and sift the n-best span candidates.

We will elaborate in next two parts (\textbf{Sentence Features} and \textbf{Span Representation}) for how to build the span representation in details.
\subsubsection{Sentence Features}
\label{sec:featuremodel}
This part describes how the sentence sequence features are built from the raw segmented sentence (Refer to the red rectangle part in Fig.~\ref{fig3} for more details). In this step, both char level and word level information are used to build the sequence features, we use the framework of ~\cite{jiecrfnetwork} to build the hidden features. Especially, we use the Conventional Neural Network (CNN)~\cite{cnnref} on character level feature building, and Long Short Term Memory Neural Network (LSTM)~\cite{lstmref} on word level feature building, which is the best combination described in the ~\cite{jiecrfnetwork}.

Moreover, before using the hidden features, we apply an attention mechanism layer over the sequence features to refine and condense the sequence hidden features. To specify, we use a pre-defined vector {$\mathbf{v_s}$} as target vector over the sequence hidden features $H = [h_1, h_2, ..., h_n]$ with dot products. Then, we make a reduce sum operation on the hidden dimension axis and compute a soft-max contribution weight on every token in the sentence. Here, ${p_i}$ is the probability (contribution weight) to the pre-defined vector {$\mathbf{v_s}$}.
\begin{equation}
    p_i = \frac{h_i \cdot \mathbf{v_s}}{\sum_{k=1}^n h_k \cdot \mathbf{v_s}} \quad (1 \leq i \leq n)
\end{equation}
The probability (contribution weight) of each token in the sentence is applied on its corresponding token. 
\begin{equation}
    h_{si} = h_i * p_i \quad (1 \leq i \leq n)
\label{eq:seqhidden}
\end{equation}
Here the $h_{si}$ can be seen as the refined feature on terminology. We use the $H_{s} = [h_{s1}, h_{s2}, ..., h_{si} ..., h_{sn}]$ as the sentence sequence features for span representation.
\subsubsection{Span Representation}
Once we get the initial candidate spans in Section~\ref{sec:termpsans} and the hidden features of the sentences in Eq.~\ref{eq:seqhidden}, we can design feature patterns and build the span representations from $H_s$ (Eq.~\ref{eq:seqhidden}).

To specify, given a sentence $S={w_1,w_2,..., w_i,...,w_n}$, its final feature layer is $H_{s} = [h_{s1}, h_{s2}, ..., h_{si} ..., h_{sn}]$. Suppose a candidate span $Span_m=[i, j], (0 \leq i \leq j \leq n \ and \  j-i \leq \textbf{\textit{k}})$, \textbf{\textit{k}} is the maximum length of the terms in the term candidate set \textbf{\textit{T}}.
\\\\
For \textbf{\textit{Each Span}}, we construct four kind features from $H_s$:\\\\
\textbf{I). \textit{Span Node}} is designed to contain the continuous information of the candidate term span for our model. For example the POS-tag sequence of the candidate. 

For a \textbf{\textit{Span}}, suppose its continuous hidden feature vectors in $H_s$ are $H_m = [h_i, h_{i+1}, ..., h_j]$, we first flatted hidden vectors and concatenate them, then we use a Multi-Layer Perceptron (MLP) to refine the the flatten vector to a vector ${V_n}$ with the same dimension on hidden features. We use the ${V_n}$ as the span node vector. 
\begin{equation}
    V_n = MLP([h_i:h_{i+1}: ...:h_j])
\end{equation}
\textbf{II). \textit{Span Head}} is designed to contain the head word information if any and whether all the words in span can form a complete Noun Phrase.

We use a pre-defined vector {$\mathbf{v_t}$} which has the same dimension with hidden features as term target vector, and apply a term attention over its hidden feature sequence $H_m$ to get its head feature vector $V_h$ of the span. 

First the vector {$\mathbf{v_t}$} is applied on $H_m$ with multiply products to reduce the hidden feature of each word to a logit.
\begin{equation}
    P_h^{[x]} = \frac{h_x * \mathbf{v_t^T}}{\sum_{x=i}^j h_k * \mathbf{v_t^T}}  \qquad (h_x, h_k \in H_m)
\end{equation}
Then the soft-max score from the logits is applied on their corresponding tokens.
\begin{equation}
    V_h = \sum_{x=i}^j h_x * P_h^{[x]} \qquad (h_x \in H_m)
\end{equation}
Here the $x$ means the token in the span token sequence.\\\\
\textbf{III). \textit{Start and End Words}} is designed to contain the feature information of begin and end word to the model. For example, generally, the term cannot start with a \textbf{PREP} word.

Given the span hidden feature sequence $H_m$, we choose the hidden feature of its start word and end word separately and concatenated them into a vector $V_{be}$.
\begin{equation}
    V_{be} = [h_i: h_j]
\end{equation}
\textbf{IV). \textit{Sentence Targeted Attention Node}} is designed to embed some feature information like whether the candidate span can express a concept to the complete sentence and leverage the information from the sentence level into term spans.

We use the mean vector of a span hidden features as a target vector and apply dot-wise attention over the sentence hidden features. This kind of attention mechanism is widely used in targeted sentiment analysis~\cite{jiangming,yuzetarget}.

To specify, given a span with hidden feature sequence $H_m$, suppose its mean vector is $\hat{h_m}$. Here $\avsum$ means 'average sum' on token axis. 
\begin{equation}
    \hat{h_m} = \avsum_{x=i}^j h_x \quad (h_x \in H_m)
\end{equation}

We use $\hat{h_m}$ as a target vector and apply its transposed on sentence level hidden feature sequence $H_{s}$ with multiply products to reduce the hidden feature of each token to a logit $\hat{h_s}$.
The logits will be softmaxed on the token axis in the sentence.
\begin{equation}
    P_s^{[x]} = \frac{h_{s}[x] * \hat{h_m}^T}{\sum_{k=1}^n h_{s}[k] * \hat{h_m}^T} \quad (h_s[x], h_s[k] \in H_s)
\end{equation}\label{eq:s}

The computed probability above is applied to its corresponding token in the sentence hiddens $H_{s}$, and the weighted sum product $V_s$ is the span targeted attention node.
\begin{equation}
    V_s = \sum_{i=1}^n h_s[x] * P_s^{[x]}
\end{equation}
Here the $\sum$ operations are applied on the token sequence axis. $H_{s}$ is the results in Eq.~\ref{eq:seqhidden}.

Apart from these four features, we also give each span a length feature vector $V_l$ to indicate the length of the span.
So, given a candidate span $Span^M$, it has a span representation $S_M$, which has a five times dimension in comparison with the hidden feature.
\begin{equation}
    S_M = [V_n, V_h, V_{be}, V_s, V_l]
\end{equation}
Here the concatenate operation over the feature dimension axis. 
\subsubsection{Additional Features*}
Other kind source features such as ELMO~\cite{elmof}\footnote{For ELMO feature, we use the pre-trained model presented by the nlp toklit allennlp: https://github.com/allenai/allennlp/blob/master/tutorials/how\_to/elmo.md}, Part-Of-Speech (POS-Tags) Embedding also can be added into the span representation. To extract as much information as possible, apart the span length feature, we will do all the four kind processing (I, II, III, IV) on every feature source.
\subsection{Classification and Ranking}
After obtaining the span representations, we use these representations do the classification and ranking steps.

First, based on the span representations $S_M$, we do a binary classification ($CLF_{FC}$) to classify these span candidates into true and false groups ($TF_G$). 
\begin{equation}
    TF_G = CLF_{FC}(S_M)
\label{eq:classifier}
\end{equation}
After classification, a scoring step will be applied over the 'true' span group ($T_G$, $T_G \in TF_G$). Here, we obtain the scores ($R_{scores}$) from a regression function($REG$) which use the span representation $S_M$ as inputs. $S_M^{T_G}$ is the span representation of true group $T_G$. The regression function($REG$) is designed to give each span candidate a score between 0 and 1.
\begin{equation}
    R_{scores} = \{REG(S_M^{T_i}),\quad S_M^{T_i} \in S_M^{T_G}\}
\end{equation}

The scores of the $T_G$ span group are then handed to a ranker as ranking (or confidence) scores. The top-K span results from the ranking step would be thought as the final output ($TM_S$) of our SCR model, which are with higher ranking scores (or confidence).
\begin{equation}
    TM_S = RANKER|_{n=1}^K(R_{scores})
\end{equation}
Here $K$ is a threshold value, we compute K = $\alpha\cdot{|TotalWords|}$. $|TotalWords|$ is the total words currently processed. $\alpha$ is a ratio indicate that how much terms are in a certain number of words. 
\subsection{Training Loss}
In our model, there are two loss source: one is the classification loss and one is the ranking step loss. We use a two-stage optimization strategy to minimizing the model loss. First, the cross-entropy loss of the classification step is computed and optimized. 
\begin{equation}
    Loss_{(classifier)} = -(y*log(p) + (1-y)*log(1-p))
\end{equation}

After getting a best result on the classifier, the parameters in the classifier are freezed, and the sigmoid-likelihood loss from the ranking step is computed and optimized. These two stage will be separately trained. We use early-stop method to select a classifier with high recall, and apply ranking on pruned space from the classifier. Moreover, the loss of the ranker is design to make the scores of true instances approach to 1 while the false's decrease to 0.
\begin{equation}
    Loss_{(ranker)} = \avsum_{{y} \in {Y_{\{gold\}}}}{(1 - Sigmoid(y))} + \avsum_{{y'} \in {Y_{\{K-gold\}}}}{Sigmoid(y')}
\end{equation}
Here, the $Y_{\{gold\}}$ are the logic regression results of the gold term instances in the restricted search space and $Y_{\{K - gold\}}$ are the results from the instances in filtered search space which does not contain gold term instances. Both $Y_{\{gold\}}$ and $Y_{\{K - gold\}}$ will be computed to a mean value over all their corresponding instances before add-up.

All the loss is optimized via the Adam optimizer with a learning rate $lr$ = 0.01. There two optimizer that correspond to the parameter set from the classifier and ranker respectively.
\section{Experiments and Analysis}
\subsection{Data}
In our experiments, we use the GENIA3.02~\cite{genia}\footnote{http://www.geniaproject.org/genia-corpus}, a human annotated, and biology corpus for information extraction and text mining systems.  

\begin{figure}\centering
\includegraphics[width=70mm,scale=0.5]{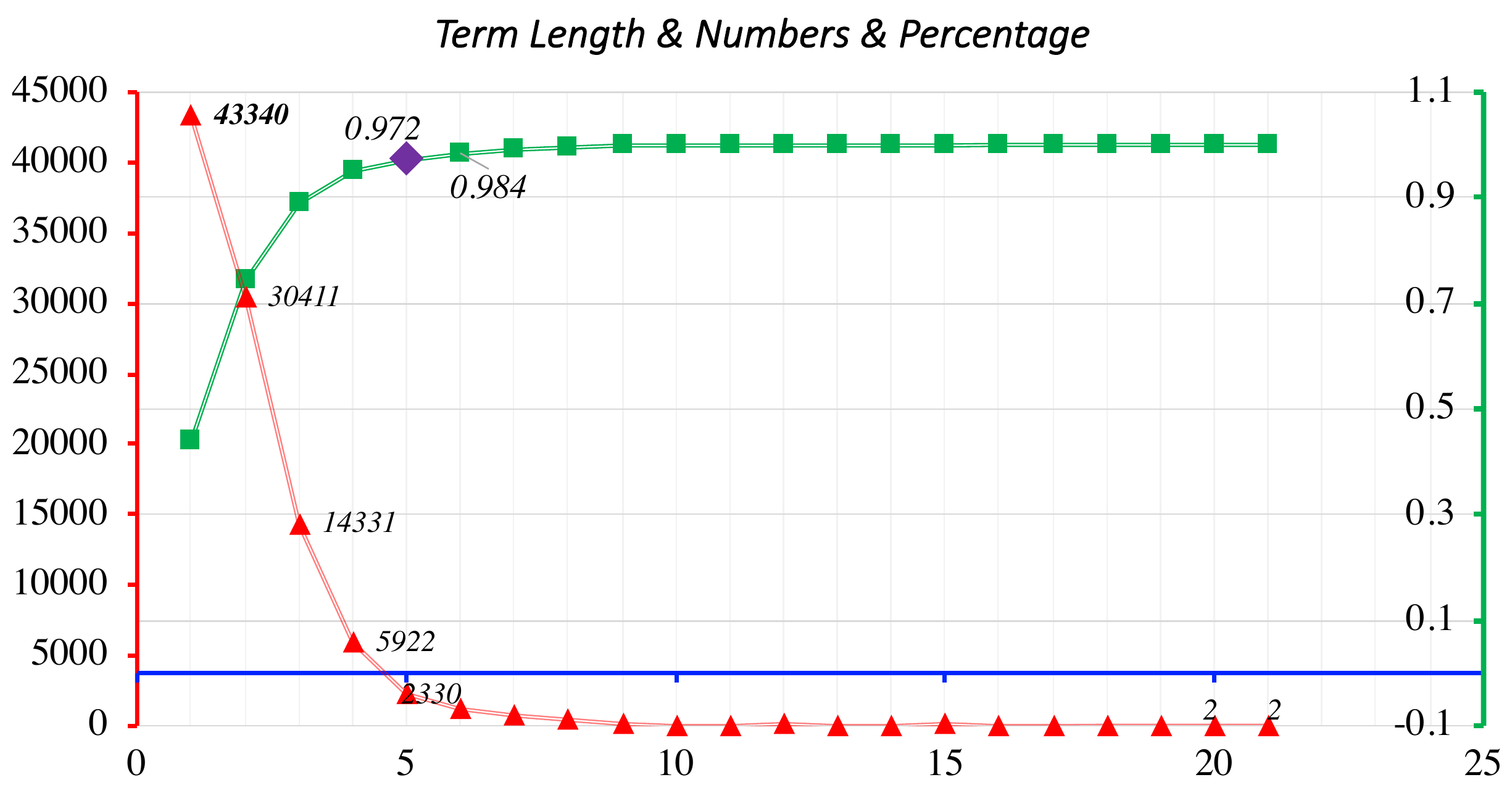} 
\caption{Term lengths \& Numbers \& Percentage distribution in the corpus}
\label{fig:termlength}
\end{figure}

An example is given in the Table~\ref{tab:terms} in Section~\ref{section:intro}, it contains nested terms. We also make a statistic on term lengths and their percentage, it is listed below in Line Figure~\ref{fig:termlength}.

There are total 99,111 terms (distribution indicated in the red line) in 18,539 sentences (total 490,766 words). In these terms, 22675 terms are nested in or overlapped with other terms. 76436 terms are independent.

The max length of the terms is 22, with most terms (97.2\%) have a length range from 1 to 5. The term ratio is 99111/490766 $\approx$ 0.202. However, the term ratio ($\alpha$) in our experiments is set to 0.23, a little bigger than true distribution to cover and recall more instances, which in turn decreases the our model precision.

We split the total corpus by sentence into Train/Dev/Test parts with a ratio 0.9 : 0.05 : 0.05 and shuffle the Train when training model. \\\hfill
\textbf{(Hyper) Parameters}  we used for our experiments are listed in the Table~\ref{tab:hyper} below.
The dropout~\cite{dropout} is only applied in the training progress to avoid over-fitting. We would stop the training processing after the evaluating loss has no increasing for a threshold times in $Early Stop$. 
\subsection{Results and Analysis}
\subsubsection{Baselines}
Currently, most ATE system are not designed to support nested term extraction. As there exists few systems supporting nested term extraction, resulting the weakness in lateral contrast. Here, we list some state-of-the-art ATE systems and their performance on the GENIA corpus.\\
1. Wang et al.~\cite{featureless}, a co-training method that uses minimal training data and achieve a comparable results with the state-of-the-art on GENIA corpus.\\
2. Yuan et al.~\cite{yuyuan}, a feature-based machine learning method using n-grams as term candidates and 10 kinds features is pre-processed for each candidate.\\
\begin{table}[!htb]
    \begin{minipage}{.45\linewidth}
        \caption{Hyper-parameters}
        \renewcommand{\arraystretch}{1.2}
        \begin{tabular}{l|l}\hline\hline
        {$DIM_{Word\ Embedding}$}&{150}\\\hline
        {$DIM_{POS-tag\ Embedding}$}&{30}\\\hline
        {$DIM_{Word\ LSTM}$}&{150}\\\hline
        {$DIM_{Span\ Length}$}&{30}\\\hline
        {Word LSTM Layers}&{2}\\\hline
        {POS-tag LSTM Layers}&{1 (optional)}\\\hline
        {Learning Rate}&{0.01}\\\hline
        {Batch Size}&{100}\\\hline
        {Random Seed}&{626}\\\hline
        {Dropout}&{0.6}\\\hline
        {Term Ratio}&{0.23}\\\hline
        {Early Stop}&{26}\\\hline
        \hline
        \end{tabular}
        \label{tab:hyper}
    \end{minipage}
    \begin{minipage}{.5\linewidth}
        \caption{Results on Test Set}
        \small
        \begin{tabular}{|c|c|c|c|c|}
        \hline
        \multicolumn{2}{|c|}{}  \normalsize & \normalsize{Precision} & Recall & F1     \\ \hline
         \multicolumn{2}{|l|}{Wang et al.~\cite{featureless}}  & {0.647}    & {0.780} & {0.707} \\ \hline
        \multicolumn{2}{|l|}{Yuan et al.~\cite{yuyuan}}  & \textbf{0.7466}       &  {0.6847}    &  {0.7143}    \\ \hline
        {} & \begin{tabular}[c]{@{}c@{}}Random\\ Embedding\end{tabular} & 0.5044 & \textbf{0.9639} & 0.6622   \\ \cline{2-5} 
                        {Our Model}     &  GloVe   & 0.5093      & \textbf{0.9557}   & 0.6575   \\ \cline{2-5} 
                        {(Classifier)}     & + POS-tag     & 0.5198      & \textbf{0.9632}   & {0.6753}   \\ \cline{2-5} 
                             & + ELMo               & 0.5220      & \textbf{0.9541}   & 0.6748   \\ \cline{2-5}
                             & + ALL                & 0.5163      & \textbf{0.9698}   & 0.6738   \\ \hline
        {} & \begin{tabular}[c]{@{}c@{}}Random\\ Embedding\end{tabular} & 0.7237 & 0.8343 & 0.7751   \\ \cline{2-5} 
                        {Our Model}     &  GloVe   & 0.7244      & 0.8356   & 0.7760   \\ \cline{2-5} 
                        {(Ranker)}     & + POS-tag     & \textbf{0.7265}      & \textbf{0.8375}   & \textbf{0.7780}   \\ \cline{2-5} 
                             & + ELMo               & 0.7252      & \textbf{0.8386}   & 0.7778   \\ \cline{2-5}
                             & + ALL                & \textbf{0.7316}      & 0.8327   & \textbf{0.7789}   \\ \hline
        \end{tabular}
        \label{tab:compare}
    \end{minipage}
    \begin{tablenotes}
   		 \item \textbf{Noted that:} Decreasing the term ratio {$\boldsymbol{\alpha}$} will increase the precision but degenerate the recall (Fig.~\ref{fig:testratio}). All the results in Tab.~\ref{tab:compare} are under the the settings in Tab.~\ref{tab:hyper}.
    \end{tablenotes}
\end{table}
Our best models for testing are chosen with the loss of development dataset and their performance is list in Tab.~\ref{tab:compare}. In the table, ~\cite{yuyuan} achieves a satisfying result with Random Forest method in their paper, but the feature preparation is complex and time-consuming. The training data is also re-balanced on the positive and negative instances.\\\\
For our classifier model, it has a high recall on the extracting terms, however the precision is not satisfying. The pre-trained word embedding [+GloVe] has little contribution, one reason is that there exists nearly 40\% of the words in dataset is out of vocabulary. With extra features like POS-tags [+POS-tag], both precision and recall increase, which indirectly gives some evidences that span representations are not precise and concrete as the external POS-tags features on the POS-tag aspect. However, the ELMO features [+ELMO], which we thought should work better than the POS-tag features, do not bring much improvement as it raises pretty little in precision and falls in recall.  When we utilize the POS-tag, ELMO feature and GloVe pre-trained embedding together [+ALL], we get a further improvement on recall of the classifier. But also cause a sharp increase in the resource consuming. The effect of ELMO feature weakens when we decrease the dimension of ELMO feature.\\\\
For the ranker model, it is expected to filter the pruned span candidates, and it gains a better precision score than the classifier, but a low recall score problem due to some loss of true positive terms. The POS-tag features [+POS-tag] bring improvement in both precision and recall but not significant than the classifier. The effect of pre-trained emdedding vector [+GloVe] also vanishes, which can be seen as a fluctuation in error margin. The ELMO features [+ELMO] do increase the recall, but not obviously. Compared with using all the features [+ALL], the ranker model uses POS-tag [+POS-tag] is slight poor in precision. \\\\
The ELMO feature does not help much to improve both our classifier and ranker model. It makes us doubt that if the 'hard' features work better than the 'soft'  features in the ATE task.
\subsection{Other Experiments}
\subsubsection{Span Length:}
We compare the model performance under different maximum span length (from 1 to 15), and list them on Test set in the Figures (Fig.~\ref{fig:testclassifier}, Fig.~\ref{fig:testranker}) on our classifier model and ranker model below.

For the classifier, as we increase the maximum term length, the recall increase to a stable value (approximate 0.96) without dropping, which means the saturation of its recall ability. It is reasonable that the precision decreases to a fluctuated point as the candidates space increases several times when increasing the maximum term length. 

For the ranker, the precision and recall increase to a stable state. However, when the length is less than 2, the ranker model has low precision. But, shorter maximum length means lower span candidates space, which means the model should obtain a higher precision. We will explain why in the next part.

Overall, it can be noticed that the final result (from the ranker) has not been influenced too much when increasing max-length, which indicating that the model will maintain stable and can distinguish the true positive instances.
\begin{figure}[!htb]
   \begin{minipage}{0.48\textwidth}
     \centering
        \begin{tikzpicture}[scale=0.75]
    	\begin{axis}[
    		height=6cm,
    		width=9cm,
    		grid=major,
    		legend style={font=\fontsize{4}{5}\selectfont,at={(0.6,0.5)}, anchor=south west,}
    	]
    	\addplot coordinates {
                            (1, 0.5956)
                            (2, 0.6044)
                            (3, 0.5567)
                            (4, 0.5354)
                            (5, 0.5031)
                            (6, 0.5163)
                            (7, 0.4266)
                            (8, 0.4190)
                            (9, 0.4506)
                            (10, 0.4028)
                            (11, 0.4300)
                            (12, 0.3928)
                            (13, 0.3553)
                            (14, 0.4037)
                            (15, 0.3894)
    	};
    	\addlegendentry{Precision}
    			
    	\addplot coordinates{
                            (1, 0.9075)
                            (2, 0.9249)
                            (3, 0.9460)
                            (4, 0.9515)
                            (5, 0.9561)
                            (6, 0.9514)
                            (7, 0.9664)
                            (8, 0.9690)
                            (9, 0.9656)
                            (10, 0.9728)
                            (11, 0.9732)
                            (12, 0.9719)
                            (13, 0.9800)
                            (14, 0.9750)
                            (15, 0.9759)
        };
    	\addlegendentry{Recall}
    			
    	\addplot coordinates{
                            (1, 0.7192)
                            (2, 0.7310)
                            (3, 0.7010)
                            (4, 0.6852)
                            (5, 0.6593)
                            (6, 0.6693)
                            (7, 0.5919)
                            (8, 0.5850)
                            (9, 0.6145)
                            (10, 0.5697)
                            (11, 0.5965)
                            (12, 0.5595)
                            (13, 0.5215)
                            (14, 0.5710)
                            (15, 0.5566)
        };
    	\addlegendentry{F1}
    	\end{axis}
    \end{tikzpicture}
     \caption{Classifier on lengths(Testset)}\label{fig:testclassifier}
   \end{minipage}\hfill
   \begin{minipage}{0.48\textwidth}
     \centering
         \begin{tikzpicture}[scale=0.75]
        	\begin{axis}[
        		height=6cm,
        		width=8cm,
        		grid=major,
        		legend style={font=\fontsize{5}{6}\selectfont,at={(0.6,0.1)}, anchor=south west,}
        	]
        	\addplot coordinates {
                                (1, 0.3265)
                                (2, 0.5932)
                                (3, 0.6880)
                                (4, 0.6954)
                                (5, 0.6980)
                                (6, 0.7201)
                                (7, 0.7109)
                                (8, 0.7193)
                                (9, 0.7235)
                                (10, 0.7078)
                                (11, 0.7119)
                                (12, 0.7211)
                                (13, 0.7071)
                                (14, 0.7199)
                                (15, 0.7175)
        	};
        	\addlegendentry{Precision}
        		
        	\addplot coordinates{
                                (1, 0.9032)
                                (2, 0.9185)
                                (3, 0.8693)
                                (4, 0.8232)
                                (5, 0.8047)
                                (6, 0.8196)
                                (7, 0.8052)
                                (8, 0.8122)
                                (9, 0.8146)
                                (10, 0.7959)
                                (11, 0.7995)
                                (12, 0.8094)
                                (13, 0.7936)
                                (14, 0.8079)
                                (15, 0.8051)
            };
        	\addlegendentry{Recall}
        	
        	\addplot coordinates{
                                (1, 0.4797)
                                (2, 0.7209)
                                (3, 0.7681)
                                (4, 0.7539)
                                (5, 0.7475)
                                (6, 0.7667)
                                (7, 0.7551)
                                (8, 0.7629)
                                (9, 0.7664)
                                (10, 0.7493)
                                (11, 0.7532)
                                (12, 0.7627)
                                (13, 0.7479)
                                (14, 0.7614)
                                (15, 0.7588)
            };
        	\addlegendentry{F1}
        	\end{axis}
        \end{tikzpicture}
     \caption{Ranker on lengths(Testset)}\label{fig:testranker}
   \end{minipage}
\end{figure}\\
\textbf{Term Ratio:} As our final model output number is controlled by a threshold $K$ which is computed by K = $\alpha\cdot{|TotalWords|}$ (the total words number is changeless), we test the ranker model performance on different term ratio $\alpha$ to see how the factor influence the final result on test set (Please refer Fig.~\ref{fig:testratio} for more details, we test the term ratio from 0.08 to 0.30). 

In the figure, the first y axis represents the Precision/Recall/F1, the second y axis represents the term span number. 
\begin{figure}[!htb]
   \begin{minipage}{0.48\textwidth}
     \centering
        \begin{tikzpicture}[scale=0.75]
     \begin{axis}[
       xmin=0.07,xmax=0.31,
       ymin=0,ymax=1,
       height=6cm,
	   width=9cm,
	   grid=major,
       axis y line*=left,
       legend style={font=\fontsize{4}{5}\selectfont,at={(0.35, 0.02)}, anchor=south west,}
     ]
       \addplot coordinates{
                            (0.08, 0.8039)
                            (0.09, 0.8003)
                            (0.1 , 0.8049)
                            (0.11, 0.8058)
                            (0.12, 0.8053)
                            (0.13, 0.8059)
                            (0.14, 0.8020)
                            (0.15, 0.7995)
                            (0.16, 0.7995)
                            (0.17, 0.7987)
                            (0.18, 0.7893)
                            (0.19, 0.7747)
                            (0.2 , 0.7512)
                            (0.21, 0.7289)
                            (0.22, 0.7065)
                            (0.23, 0.6874)
                            (0.24, 0.6677)
                            (0.25, 0.6497)
                            (0.26, 0.6313)
                            (0.27, 0.6137)
                            (0.28, 0.5986)
                            (0.29, 0.5864)
                            (0.3 , 0.5750)
       };\addlegendentry{Precision}
      \addplot coordinates{
                            (0.08, 0.3176)
                            (0.09, 0.3558)
                            (0.1 , 0.3978)
                            (0.11, 0.4382)
                            (0.12, 0.4778)
                            (0.13, 0.5179)
                            (0.14, 0.5551)
                            (0.15, 0.5930)
                            (0.16, 0.6326)
                            (0.17, 0.6717)
                            (0.18, 0.7027)
                            (0.19, 0.7281)
                            (0.2 , 0.7433)
                            (0.21, 0.7572)
                            (0.22, 0.7687)
                            (0.23, 0.7824)
                            (0.24, 0.7931)
                            (0.25, 0.8036)
                            (0.26, 0.8122)
                            (0.27, 0.8200)
                            (0.28, 0.8294)
                            (0.29, 0.8415)
                            (0.3 , 0.8536)
       };\addlegendentry{Recall}
      \addplot coordinates{
                            (0.08, 0.4553)
                            (0.09, 0.4926)
                            (0.1 , 0.5325)
                            (0.11, 0.5677)
                            (0.12, 0.5998)
                            (0.13, 0.6305)
                            (0.14, 0.6561)
                            (0.15, 0.6809)
                            (0.16, 0.7063)
                            (0.17, 0.7297)
                            (0.18, 0.7435)
                            (0.19, 0.7507)
                            (0.2 , 0.7473)
                            (0.21, 0.7427)
                            (0.22, 0.7363)
                            (0.23, 0.7318)
                            (0.24, 0.7250)
                            (0.25, 0.7185)
                            (0.26, 0.7104)
                            (0.27, 0.7020)
                            (0.28, 0.6954)
                            (0.29, 0.6911)
                            (0.3 , 0.6871)
       };\addlegendentry{F1}
     \end{axis}
     \begin{axis}[
       xmin = 0.07, xmax = 0.31,
       ymin = 1000, ymax = 8000,
       hide x axis,
       height=6cm,
	   width=9cm,
	   grid=major,
       axis y line*=right,
       legend style={font=\fontsize{4}{5}\selectfont,at={(0.63, 0.02)}, anchor=south west,}
     ]
       \addplot [purple, thick] coordinates{
                            (0.08, 2024)
                            (0.09, 2278)
                            (0.1 , 2532)
                            (0.11, 2786)
                            (0.12, 3040)
                            (0.13, 3292)
                            (0.14, 3546)
                            (0.15, 3800)
                            (0.16, 4054)
                            (0.17, 4308)
                            (0.18, 4561)
                            (0.19, 4815)
                            (0.2 , 5069)
                            (0.21, 5322)
                            (0.22, 5574)
                            (0.23, 5831)
                            (0.24, 6085)
                            (0.25, 6337)
                            (0.26, 6591)
                            (0.27, 6845)
                            (0.28, 7098)
                            (0.29, 7352)
                            (0.3 , 7605)
       };  \addlegendentry{K-num}
       
       \addplot [cyan, thick] coordinates{
                            (0.08, 5123)
                            (0.09, 5123)
                            (0.1 , 5123)
                            (0.11, 5123)
                            (0.12, 5123)
                            (0.13, 5123)
                            (0.14, 5123)
                            (0.15, 5123)
                            (0.16, 5123)
                            (0.17, 5123)
                            (0.18, 5123)
                            (0.19, 5123)
                            (0.2 , 5123)
                            (0.21, 5123)
                            (0.22, 5123)
                            (0.23, 5123)
                            (0.24, 5123)
                            (0.25, 5123)
                            (0.26, 5123)
                            (0.27, 5123)
                            (0.28, 5123)
                            (0.29, 5123)
                            (0.3 , 5123)
       }; \addlegendentry{True-Term-num}

       \addplot [green, thick] plot [smooth] coordinates{
                            (0.08, 1627)
                            (0.09, 1823)
                            (0.1 , 2038)
                            (0.11, 2245)
                            (0.12, 2448)
                            (0.13, 2653)
                            (0.14, 2844)
                            (0.15, 3038)
                            (0.16, 3241)
                            (0.17, 3441)
                            (0.18, 3600)
                            (0.19, 3730)
                            (0.2 , 3808)
                            (0.21, 3879)
                            (0.22, 3938)
                            (0.23, 4008)
                            (0.24, 4063)
                            (0.25, 4117)
                            (0.26, 4161)
                            (0.27, 4201)
                            (0.28, 4249)
                            (0.29, 4311)
                            (0.3 , 4373)
       }; \addlegendentry{True Positive}
     \end{axis}
   \end{tikzpicture}
     \caption{Ranker on Term Ratios(Testset)}\label{fig:testratio}
   \end{minipage}\hfill
   \begin{minipage}{0.48\textwidth}
     \centering
         \begin{tikzpicture}[scale=0.75]
        \begin{axis}[
                xmin=0.09,
                xmax=0.32,
                ymin=0,
                ymax=230,
        		height=6cm,
        		width=7.5cm,
        		grid=major,]
        \addplot [black, fill=red!20] plot [smooth] 
            coordinates {
            (0.09, 196)
            (0.1 , 215)
            (0.11, 207)
            (0.12, 203)
            (0.13, 205)
            (0.14, 191)
            (0.15, 194)
            (0.16, 193)
            (0.17, 180)
            (0.18, 159)
            (0.19, 130)
            (0.2 , 78)
            (0.21, 71)
            (0.22, 59)
            (0.23, 70)
            (0.24, 55)
            (0.25, 54)
            (0.26, 48)
            (0.27, 46)
            (0.28, 48)
            (0.29, 54)
            (0.3 , 45)}
             |- (0.09,0) |- (0.3,0) -- cycle;
        \addplot [red, thick] (0.199,0) |- (0.19,300);
        \end{axis}

        \end{tikzpicture}
     \caption{True Positive Terms Distribution}\label{fig:testdis}
   \end{minipage}
\end{figure}\\
The recall {\color{red}[Red Line]} increases gradually with the increasing term ratio, and also the precision {\color{blue}[Blue Line]} decreases due to the increasing of candidates space {\color{purple}[Purple Line, K-num]}. When the ratio approach 0.2 (the actual term distribution ratio in the dataset, the True-Term-num {\color{cyan}[Cyan Line]} and {\color{purple} K-num} crossing point), the F1 value also achieve the best. The True Positive {\color{green}[Green Line]} means the true positive instance number recalled by the ranker model.

So, in the \textbf{Span Length}, the ranker model achieve a low score when the span length equal 1 or 2 due to the large {\color{purple} K-num}, we set the ratio threshold (0.23, about 5700 candidates in test set) to run all experiments, while the actual term number (ratio) at length equal 1 and 2 in test set is 2254(0.09) and 3995(0.16), increasing 3450 and 1700 unsure. This is a disadvantage of the threshold-based ranking method. But the precision of the ranker model can be increased by decreasing the term ratio.

\begin{figure}
\centering
\includegraphics[width=125mm,scale=1]{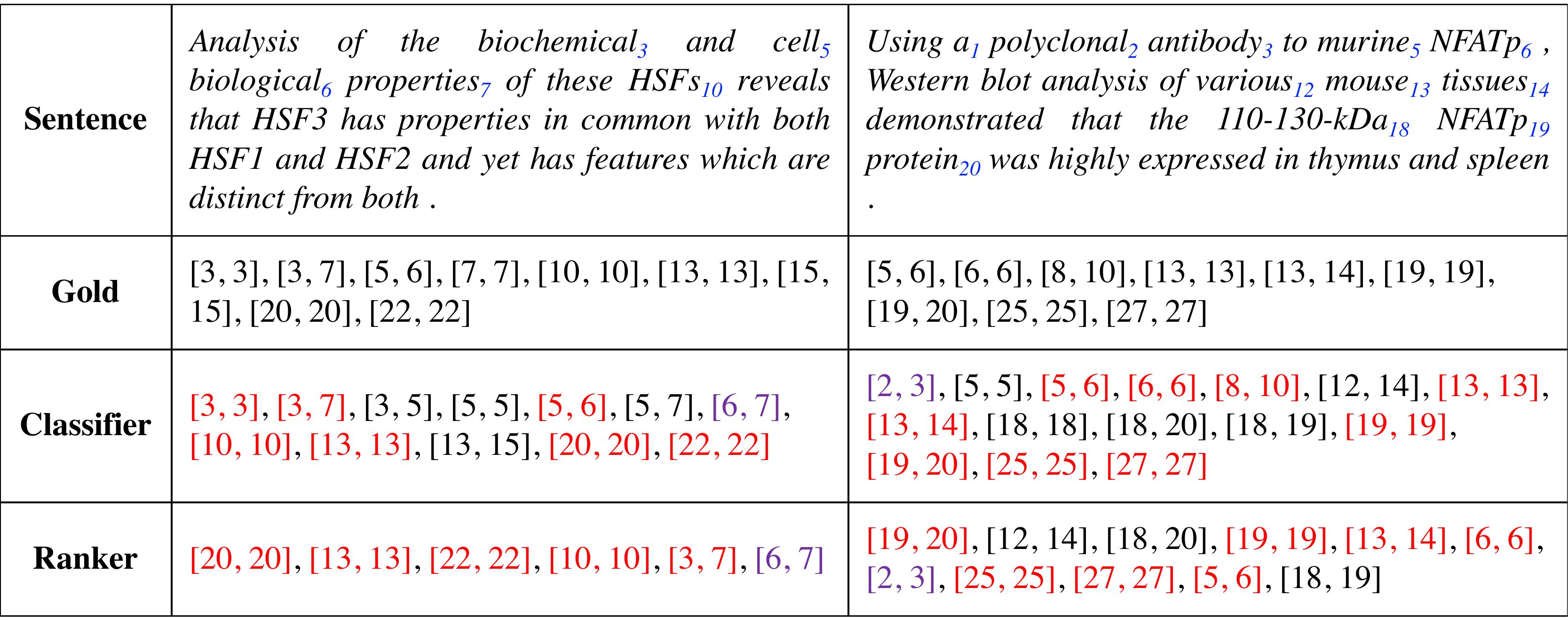}
\caption{Samples on Test-Set} 
\label{fig:sample_test}
\end{figure}

We also analyze the predicted true positive term distribution when we set the term ratio as 0.3, Fig.~\ref{fig:testdis} shows the distribution from 0.09 to 0.30, the figure indicates the term number (axis $y$) at ratio (axis $x$) in the (term ratio=0.3) outputs. The ranking scores concentrate the candidates and move them forward in the whole searching space. \\\\ 
\textbf{Samples:} We list two examples on Test-set and put them in Fig.~\ref{fig:sample_test} (the ranker results are in score decreasing order). Some predicted spans are potential to convey a term concept. In the figure, [6, 7] in the first sample and [2, 3] in the second sample are not in the gold, but they have some conceptual meanings.
\section{Conclusion and Future Work}
We proposed a deep learning-based end-to-end term extraction method in this paper. It employs classification and ranking on the span (n-grams) candidates in the sentences. Compared with current methods, it supports the nested term extraction and can achieve a comparable result with merely the segmented raw text as input. 

Based on the sentence and term span hidden features, four kinds  reasonable feature patterns are designed to convey different information. Experimental results show that these features indeed can embed some information. Though the model achieve a satisfying results, the ranker still loss many true positive instances, which decrease the recall score from 0.95 to 0.83. Moreover, threshold-based output is not so applicable on unknown or unfamiliar domain or data as we do not know the term distribution and ratio.

Future works may focus on is to immigrate the architecture on a better feature extracting model (such as BERT~\cite{bert} or GPT2.0~\cite{gpt2}). The ranking step should be degisned more reasonable to pick up the outputs. More reasonable feature patterns can be designed to convey useful information.
\section*{Acknowledgement}
Thanks for the detailed comments and suggestions on an earlier draft from three anonymous reviewers.

\end{document}